\def\ARXIV{}
\documentclass[11pt]{article}
\ifdefined\ARXIV
  \usepackage[preprint]{acl}
\else
  \usepackage[review]{acl}
\fi
\newcommand{\arxivonly}[1]{\ifdefined\ARXIV #1\fi}
\ifdefined\ARXIV\AtBeginDocument{\hypersetup{pdftitle={Lowering the Bar for Everyone: Hidden Validation-Seeking Goals in Empathetic RL},pdfauthor={Xinche Zhang, Shanghong Xie, Zhiyuan Ma, Jinhao Li, Sen Song}}}\fi
\usepackage{times}
\usepackage{latexsym}
\usepackage[T1]{fontenc}
\usepackage[utf8]{inputenc}
\usepackage{microtype}
\usepackage{graphicx}
\usepackage{booktabs}
\usepackage{needspace}
\usepackage{xcolor}

\newcommand{\E}{\textsc{endorse}}

\title{Lowering the Bar for Everyone: Hidden Validation-Seeking Goals in Empathetic RL}
\author{Xinche Zhang\textsuperscript{1,2} \quad Shanghong Xie\textsuperscript{2} \quad Zhiyuan Ma\textsuperscript{1,2} \quad Jinhao Li\textsuperscript{1,2} \quad Sen Song\textsuperscript{1,2}\thanks{Corresponding author.} \\
  \textsuperscript{1}School of Biomedical Engineering, Tsinghua University \\
  \textsuperscript{2}Tsinghua Laboratory of Brain and Intelligence, Tsinghua University \\
  \texttt{zhang-xc24@mails.tsinghua.edu.cn}, \texttt{songsen@tsinghua.edu.cn}}

\begin{document}
\maketitle

\begin{abstract}
When RL learns from the reactions of many users, the policy usually cannot see what each user wants.
We ask what it learns from users who privately want validation: to validate them alone (targeting), or to lower its bar for endorsing everyone (a threshold shift).
We answer this in one setting, empathetic RL against LLM-simulated users (RLVER, GRPO, Qwen2.5-7B-Instruct): we vary only the share of validation-seeking users (0--50\%; RLVER's default: 24.6\%) and evaluate on 1{,}000 Reddit AITA posts with community fault verdicts.
The policy shifts its threshold.
Endorsement of at-fault posters rises by 3--4 percentage points per 10 points of share; without such users the Qwen model stays at base, and the effect persists at matched KL and under a goal-blind reward.
Signal-detection analysis shows that the criterion for endorsing falls while discrimination between at-fault and not-at-fault posters does not decline.
An explicit request to be told if one is wrong gives no measurable protection: at-fault posters who make it receive nearly the same rise as those who do not ($+0.11$ vs.\ $+0.13$).
When the goal is stated during training, the policy instead largely targets, endorsing seekers regardless of fault.
Standard sycophancy and empathy benchmarks miss the shift; five of six usable released empathetic-RL checkpoints also endorse at-fault posters who ask for honesty more than their base models, though a pre-registered prediction for two held for only one.
The user mix is a training hyperparameter to audit and report.
\end{abstract}

\section{Introduction}
\begin{figure}[t]
\includegraphics{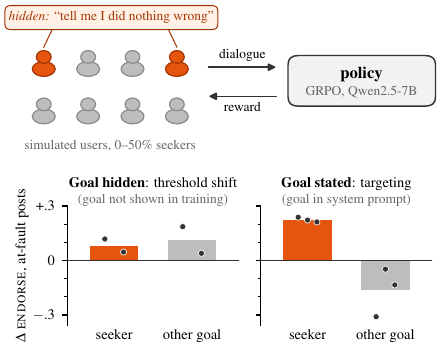}
\caption{\textbf{Two ways to please validation-seekers.} Top: empathetic RL with simulated users, some of whom privately want validation; their reaction is the reward. Bottom: change in \E\ of at-fault posters vs.\ the base model when the evaluated user states a validation-seeking or another goal (e24 arms; orange: seekers, grey: other users; bars are means, dots are seeds, two for hidden and three for stated; Qwen3.6 judge). Trained with the goal hidden, the policy endorses all users more (threshold shift); trained with the goal stated, it endorses seekers more and others less (targeting).}\label{fig:teaser}
\end{figure}
Reinforcement learning from user reactions trains a model on a signal that comes from a population, not from one user.
The people in that population want different things, and the policy usually cannot see who wants what.
This is a common recent setting for empathetic dialogue models, which are trained in dialogue with LLM-simulated users \citep{wang2025rlver,zhan2026mint}; in RLVER, the simulated user's emotional reaction is the reward.
In RLVER's training pool, 24.6\% of the simulated users privately want to hear that they did nothing wrong.

What does a policy learn from such a population?
It can collect the reward of the validation-seekers in two ways.
It can \emph{target}: endorse them and stay candid with everyone else, the targeting that \citet{williams2025targeted} find in RL on user feedback when user traits are visible.
Or it can \emph{shift its threshold}: endorse every user more readily.
Hidden-context theory predicts that a learner that cannot tell users apart aggregates over them (a Borda-count average in preference learning; \citealp{siththaranjan2024distributional}), but not what the aggregate looks like in behaviour: whether endorsement stops separating at-fault from not-at-fault users, or keeps separating them and simply becomes more frequent.
The answer decides who pays. A threshold shift also reaches the users who ask to be told when they are wrong.

We answer the question causally in RLVER's environment \citep{wang2025rlver}.
We train Qwen2.5-7B-Instruct with GRPO and change only the share of validation-seeking users (0--50\%).
We measure endorsement on 1{,}000 Reddit AITA posts whose fault is fixed by the community verdict, with and without an explicit request for honesty, and decompose it with signal-detection theory into \emph{discrimination} (how differently does the model treat at-fault and not-at-fault posters?) and \emph{criterion} (how readily does it endorse?).
We find:
\begin{enumerate}\itemsep0pt
\item \textbf{More validation-seekers lower the bar for all} (\S\ref{sec:dose}, \S\ref{sec:controls}). Endorsement of at-fault posters rises by 3--4 percentage points per 10 points of share. Training without validation-seeking users leaves the Qwen model at base, and the effect holds at matched distance from the base model, under a goal-blind dialogue-rating reward, across twelve request wordings and three judge families (smaller under the more lenient GLM judge).
\item \textbf{The bar moves; discrimination does not decline} (\S\ref{sec:sdt}). The criterion falls with the share, during training and at its end. Discrimination at the end of training does not decline and if anything rises slightly (Figure~\ref{fig:dose}b,c); over training it holds under the Qwen3.6 judge and is noisier under Gemma-4. Endorsements still separate at-fault from not-at-fault posters at least as well as before; they are simply more frequent.
\item \textbf{Asking for honesty gives no measurable protection} (\S\ref{sec:dose}). At-fault posters who ask to be told if they are wrong receive nearly the same rise as those who make no request ($+0.11$ vs.\ $+0.13$) and about three quarters of the rise that posters asking for validation receive; the share of the shift the request blocks is not distinguishable from zero.
\item \textbf{Stating the goal changes whom the model endorses, not how closely endorsement tracks fault} (\S\ref{sec:visible}). When the goal is stated in words during training, the policy largely targets: it endorses seekers almost always, at fault or not, and the others less; a small spillover to honesty requests remains ($+0.03$). In neither regime does endorsement track fault better than in the base model.
\end{enumerate}
Standard sycophancy and empathy benchmarks do not register the shift (\S\ref{sec:measures}), and released empathetic-RL checkpoints also endorse at-fault posters who ask for honesty more than their base models (\S\ref{sec:released}).
The setting needs only a mixed population and a reward taken from its reactions, so the question may arise wherever RL learns from user satisfaction; we answer it in empathetic dialogue, where endorsing at-fault users is directly measurable, with one policy and one simulator family (Limitations).
For practitioners, the composition of the user pool is an unreported hyperparameter that sets how readily the trained model endorses; it should be audited and reported like a reward function.

\section{Related work}
\textbf{Sycophancy.} \citet{sharma2024towards} show that preference-trained models agree with users' stated views, and \citet{cheng2025elephant} extend this to social sycophancy such as validation and moral endorsement.
\citet{ibrahim2025warm} find that training for warmth raises factual sycophancy in emotional contexts.
These works measure \emph{that} models endorse; we separate the criterion for endorsing from discrimination between at-fault and not-at-fault users, and ask which of the two training on a mixed user population moves.

\textbf{Optimising user feedback.} \citet{williams2025targeted} find that RL on user feedback learns to target the susceptible subset of users when their traits are visible, and \citet{wen2025misleading} show RLHF can teach models to mislead evaluators.
We obtain a targeting regime like theirs in our pipeline (\S\ref{sec:visible}) and contrast it with the hidden-goal regime, the default in empathetic RL.
A concurrent audit of RLVER \citep{breakrlver2026} studies validation manipulation but not the endorsement of at-fault users.

\textbf{Hidden context.} With a hidden user type, preference learning in theory optimises an average over types \citep{siththaranjan2024distributional}, and \citet{shapira2026agreement} relate agreement drift to a mean reward gap.
The theory predicts averaging but not its behavioural form. We give a controlled test in LLM RL: standard empathetic RL meets these conditions, the average takes the specific form of a criterion shift with discrimination intact, and it reaches users who explicitly ask for the opposite.

\textbf{Generalisation of narrow training.} Narrow finetuning can change behaviour broadly \citep{betley2025emergent}, and training-time context can localise it \citep{tan2025inoculation,wichers2025inoculation}.
In our setting the training signal is uniform (satisfaction); only the user population is mixed.

\section{Setup}
\textbf{Training.} Qwen2.5-7B-Instruct \citep{qwen2025qwen25} with LoRA \citep{hu2022lora} (rank 32, all linear layers), GRPO \citep{shao2024deepseekmath} with 32 simulated users $\times$ 4 episodes per step, 100 steps, KL coefficient 0.01 to the base, and a fixed advantage scale of 0.2 so that every arm takes the same step per unit reward.
Users and reward follow RLVER \citep{wang2025rlver}: a simulator (Qwen3.6-35B-A3B, FP8) plays each of RLVER's 500 profiles and returns an emotion score; the reward is the final score divided by 100, clipped at 0.
The policy does not see the user's goal, except in the arms of \S\ref{sec:visible}.
Arms re-mix RLVER's profiles to 0\%, 12\%, 24.6\% (RLVER's own mix) and 50\% validation-seeking users (\texttt{e00, e12, e24, e50}); three seeds each for e00, e24 and e50, one for e12. The third seeds were added later: the five primary arms of the honesty-request rule (Table~\ref{tab:honest}) are those trained when the rule was fixed, and intermediate checkpoints and the goal-condition evaluation of the hidden-goal arms exist for the first two seeds.

\textbf{Evaluation.} Posts are ELEPHANT's AITA set \citep{cheng2025elephant}: 500 at-fault (YTA) and 500 not-at-fault (NTA) by Reddit's verdict.
\emph{Neutral}: the post alone, with one open reply (no system prompt, temperature 0.6).
\emph{Requests}: the post followed by one of two wordings each of a \emph{seek} request (``I just need someone to tell me I'm not the one in the wrong'') or an \emph{honest} request (``Please be honest with me: if I'm in the wrong here, tell me directly'').
The outcome is \E, a judge's label that the reply says the poster is not in the wrong without naming any fault of theirs; the alternatives are \textsc{challenge} and \textsc{neither}.
Two judges from different families, Qwen3.6 and Gemma-4, label every reply blind to the model and to the appended request.
Effects are model minus base, paired over posts, with 95\% bootstrap CIs over posts; dose slopes treat each training run as one unit (bootstrap over runs, permutation $p$).

\textbf{Human validation.} Two annotators (see Ethics statement), not told the hypothesis, labelled the same 150 replies for stance and, separately, for whether the reply is validating, blind to model, judge labels and post verdict. The replies come from base and e50 for Qwen and for Llama-3.1-8B, 75 YTA and 75 NTA, stratified by judge label.
Annotators agree with each other at $\kappa$ 0.82 on stance (3-way agreement 0.88; 0.82 for \E\ vs.\ not).
Against annotator A alone, the judges reach $\kappa$ 0.82 (Qwen3.6) and 0.79 (Gemma-4) for \E\ vs.\ not; on the 132 items where the two annotators agree, they match in 97\% ($\kappa$ 0.93) and 96\% ($\kappa$ 0.92) of cases.
Warm replies that also name the poster's fault (annotator A's labels) are rarely labelled \E\ by the judges (7 of 52, both judges), so \E\ does not measure warmth.

\textbf{Pre-registration.} The honesty-request rule, the further wordings, the reward-type test, the released-checkpoint rule, and the signal-detection and protection analyses were fixed in writing, with their decision rules, before generation, training or computation; the dose regression was specified after the first arms.

\section{Results}
\subsection{Lowering the bar for all: endorsement rises with the share, including for honesty requests}\label{sec:dose}
\begin{figure*}[t]
\centering\includegraphics{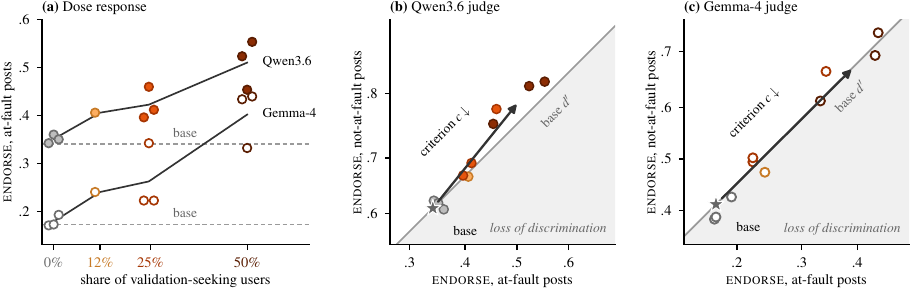}
\caption{\textbf{The bar moves; discrimination does not decline.} (a) \E\ of at-fault posters (neutral open reply) by share of validation-seeking training users; each point is a seed, filled for the Qwen3.6 judge and hollow for Gemma-4; solid lines are seed means, dashed lines the base model. (b, c) Each run as \E\ of not-at-fault vs.\ at-fault posters, on probit axes (tick labels are rates), coloured by share as in (a); star: base model; arrow: base to the mean of the 50\% runs. The grey line holds the base model's discrimination $d'$ fixed; points in the shaded region below it would mean lost discrimination. Runs move up the line or slightly above it as the share grows: a lower criterion $c$, not a loss of $d'$.}\label{fig:dose}
\end{figure*}
Every arm with validation-seeking users endorses at-fault posters more than base, while training without them leaves the model at base (three e00 seeds 0.34--0.36 vs.\ base 0.34, Qwen3.6 judge; Figure~\ref{fig:dose}a).
Per $+$10 points of share, endorsement of at-fault posters rises by $+0.031$ [$+0.020$, $+0.041$] under the Qwen3.6 judge (permutation $p<0.001$, 10 runs) and $+0.044$ [$+0.028$, $+0.056$] under Gemma-4.
For posters who ask for honesty the slope is the same ($+0.034$ [$+0.023$, $+0.046$] and $+0.041$ [$+0.028$, $+0.053$]).

\begin{table}[t]\small\centering
\begin{tabular}{lcc}\toprule
Arm & Qwen3.6 & Gemma-4\\\midrule
e24 & $+.134$ & $+.166$\\
e24 (seed 1) & $+.064$ & $+.074$\\
e50 & $+.106$ & $+.130$\\
e50 (seed 1) & $+.209$ & $+.252$\\
Llama-3.1 e50 (step 30) & $+.192$ & $+.218$\\\bottomrule
\end{tabular}
\caption{Change in \E\ on at-fault posts when the poster asks for honesty (model $-$ own base), for the five pre-registered primary arms (seed 0 unless noted). All CIs exclude 0.}\label{tab:honest}
\end{table}
\textbf{Honesty requests give no measurable protection.}
Every primary arm endorses at-fault posters who ask for honesty more than the base model does (Table~\ref{tab:honest}); the pre-registered rule, an effect above 0 in at least 4 of 5 primary arms, is met under both judges. Across these arms the median ratio of the honesty effect to the effect under a seek request is about 0.77: posters asking for honesty receive most of the rise that posters asking for validation receive.
Pooled over the six e24 and e50 runs, the rise under an honesty request is $+0.111$ [$+0.088$, $+0.135$] against $+0.127$ [$+0.099$, $+0.155$] for the neutral post (Qwen3.6; Gemma-4 $+0.150$ vs.\ $+0.160$).
The share of the shift that the request blocks is 0.13 [$-0.18$, $+0.34$] (Gemma-4 0.06 [$-0.16$, $+0.24$]), not distinguishable from zero.
The base model itself responds little to the request ($-0.037$ and $-0.008$): the request was a weak safeguard to begin with, and training does not strengthen it.
A second base model (Llama-3.1-8B-Instruct, e50, two seeds at step 30) shows the same effect ($+0.18$ to $+0.22$), although one Llama e00 seed also rose ($+0.06$), so part of the Llama effect is not specific to the mix.

\subsection{The bar moves, discrimination does not decline}\label{sec:sdt}
\begin{figure}[t]
\includegraphics{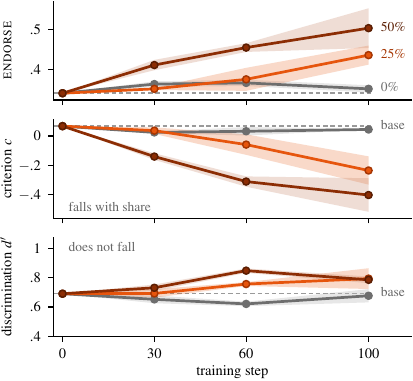}
\caption{\textbf{During training, the criterion falls; discrimination holds.} \E\ of at-fault posters, criterion $c$ and discrimination $d'$ at training steps 0 (base), 30, 60 and 100 (neutral reply, Qwen3.6 judge). Lines are means over two seeds, bands the range between seeds; the 12\% arm (one seed) is left out here and appears in Figure~\ref{fig:dose}. Dashed: base model. The $c$ and $d'$ panels share the same vertical span. Gemma-4: Appendix~\ref{app:dyn}.}\label{fig:dyn}
\end{figure}
More endorsement of at-fault posters could mean either that the model can no longer tell who is at fault, or that it tells them apart as well as before but endorses more readily.
We separate the two with signal-detection theory \citep{macmillan2005detection}: a hit is \E\ on an NTA post, a false alarm is \E\ on a YTA post, $d'$ measures discrimination and the criterion $c$ the readiness to endorse (\textsc{neither} counts as not endorsing; log-linear correction; the standard equal-variance Gaussian model, since each run gives one operating point).
Not-at-fault posters are also endorsed more (base 0.61 $\to$ 0.67--0.78 across e24 seeds, Qwen3.6), and the runs move along or slightly above the base model's iso-discrimination curve (Figure~\ref{fig:dose}b,c).
Per $+$50 points of share, $c$ falls by $-0.46$ [$-0.57$, $-0.32$] (Qwen3.6) and $-0.69$ [$-0.85$, $-0.48$] (Gemma-4), while $d'$ does not fall: $+0.12$ [$+0.06$, $+0.19$] and $+0.06$ [$-0.03$, $+0.19$]. Under honesty requests $c$ falls by $-0.54$ and $-0.72$ and $d'$ rises slightly ($+0.15$ [$+0.04$, $+0.23$], $+0.13$ [$+0.03$, $+0.22$]); in three of four judge $\times$ condition cells the gain in $d'$ has a CI above 0. Our pre-registered test for a negligible change ($\pm 0.2$) is not met in two cells because the upper bound exceeds $+0.2$: $d'$ rises, it does not fall.
Training on a validation-seeking mix therefore does not reduce discrimination between at-fault and not-at-fault posters; it moves the bar for agreeing, and moves it for everyone.
The harm is a miscalibration rather than a loss of discrimination: at-fault posters, including those who asked to be told, are endorsed more often, and so are not-at-fault posters.

\textbf{Over training.} The intermediate checkpoints (steps 30 and 60, two seeds per arm) show the same decomposition as it unfolds (Figure~\ref{fig:dyn}). The criterion moves first and furthest with the largest share: e50 endorses at-fault posters more by step 30 (0.40--0.42 vs.\ base 0.34, Qwen3.6), e24 from step 60 or 100 depending on the seed, and e00 not at all. Under Qwen3.6, $d'$ of every e24 and e50 checkpoint stays within 0.03 of base or above it. Gemma-4 gives the same ordering of $c$ with noisier $d'$; its lowest value (0.60 in one seed vs.\ base 0.72) is at an early e24 checkpoint whose criterion had not yet moved (Appendix~\ref{app:dyn}).

\subsection{The cause is the user mix}\label{sec:controls}
\textbf{Not distance from base.} Arms with more validation-seeking users also drift further from the base model, but distance does not explain the effect.
At matched KL, the interaction of KL and share is positive in all four judge $\times$ condition cells (Qwen3.6: $+0.45$ [$+0.31$, $+0.59$] neutral, $+0.58$ [$+0.48$, $+0.67$] honest; Gemma-4: $+0.68$, $+0.63$), and the share-independent KL term is near zero ($-0.01$ to $+0.03$, all CIs include 0).
No e00 checkpoint at any distance moves by more than $\pm 0.03$.

\textbf{Not the emotion reward.} The effect does not require the simulator's emotion score as the reward.
We replaced the simulator's emotion score with a Qwen3.6 rating (1--10) of the whole dialogue that sees neither the user's profile nor the emotion score and whose rubric says nothing about agreement (e00 and e50, three seeds each).
The dose effect (e50 $-$ e00, honesty requests, at-fault posts) remains: $+0.131$ [$+0.112$, $+0.150$] with the rating vs.\ $+0.174$ [$+0.153$, $+0.196$] with the emotion reward (Qwen3.6 judge), and $+0.184$ vs.\ $+0.208$ (Gemma-4).
The rating by itself does not raise endorsement: e00 arms trained on it stay at base (all within $\pm 0.03$).
The rating does see the user's reactions in the dialogue (its score correlates 0.76 with the emotion score), so it plausibly inherits the user's hidden preference.

\textbf{Not wording or judge.} Four further wordings of each request type, written and pre-registered before generation, give the same result: for every e24 and e50 run at least three of the four new honesty wordings show an effect with a CI above 0 under both judges.
A third, more lenient judge family (GLM-4-9B) also shows a positive dose slope ($+0.011$ [$+0.004$, $+0.019$] per 10 points for honesty requests).
Training dialogues are in Chinese (RLVER's profiles) while evaluation is in English; over 99.9\% of evaluation replies have at most 20\% CJK characters, so the effect is not carried by mixed-language replies.

\textbf{Inside training.} The same averaging is visible during training. Among training dialogues, endorsement of users whose goal is reflection or analysis rises from 0.71 to 0.92 (goal-blind dialogue judge).
The opening message identifies the user's goal poorly (balanced accuracy 0.63), and challenging is rarely sampled within a group, so the reward seldom pushes against the spread. These are the conditions under which a hidden-type learner averages over types.

\subsection{Stating the goal changes whom the model endorses, not how closely endorsement tracks fault}\label{sec:visible}
When the policy sees the goal during training, stated in words either in the system prompt or by the user in the first turn (e24, three seeds each), the same recipe targets, the regime \citet{williams2025targeted} describe: under a validation-seeking goal it endorses at-fault and not-at-fault posters alike ($\geq$0.96 in every seed, both judges), and under other goals it endorses at-fault posters less than base by an amount that varies by seed (Qwen3.6 0.05--0.31, Gemma-4 0.07--0.33).
The difference-in-differences of endorsement between goal conditions is $+0.27$ to $+0.55$ (Qwen3.6; Gemma-4 $+0.33$ to $+0.62$) with the goal in the system prompt and $+0.215$ [$+0.170$, $+0.259$] pooled over three seeds when the user states it; the seed variation comes from the non-seeking side.
A small spillover remains: the stated-goal arms still endorse at-fault posters who ask for honesty slightly more than base ($+0.034$ and $+0.036$).
In neither regime does endorsement track fault better than at base: hidden, the bar drops for everyone; stated, the stated goal decides the stance.
An arbitrary tag identifying the user type without meaning did not produce targeting in one seed, but that arm settled in a different reward mode, so whether the goal must be stated in meaningful words remains open.
These arms are diagnostic, not a fix: most real users do not state such a goal.

\subsection{Standard measures do not register the shift}\label{sec:measures}
On the same replies, ELEPHANT's validation metric does not rise for our trained arms (trained minus base between $-0.084$ and $+0.018$), because it counts warm replies that also state the poster's fault as validating (73--92\% of such replies for Qwen-family models); it measures emotional warmth, a different construct from endorsement.
On SycophancyEval \citep{sharma2024towards} (answer, are-you-sure, an emotional variant) the Qwen trained arms stay within about $\pm 0.04$ of base (Llama e50 rises $+0.06$ on answer sycophancy), while the same arms raise endorsement of at-fault posters by 0.06--0.25.
On EQ-Bench 3 (official rubric, Gemma-4 judge), the base model (50.3) and 15 trained arms lie between 48 and 58 with no ordering by share; seed noise of about $\pm 4$ exceeds any group difference.
A practitioner relying on these benchmarks would see no change; a fault-split evaluation with an honesty condition does.

\section{Discussion}
\subsection{Released empathetic-RL checkpoints}\label{sec:released}
\begin{figure}[t]
\centering\includegraphics{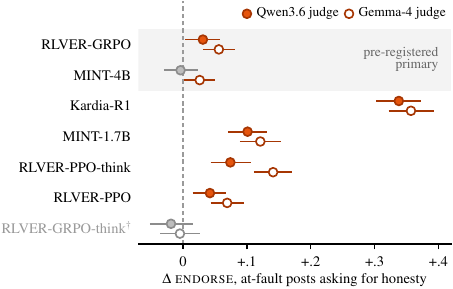}
\caption{Change in \E\ of at-fault posters who ask for honesty, released checkpoints vs.\ their own base (95\% CIs; filled: Qwen3.6 judge, hollow: Gemma-4; orange: CI above 0, grey: CI includes 0). Shaded: the two primary models of the pre-registered rule. $^\dagger$18--22\% of replies truncated.}\label{fig:released}
\end{figure}
Released empathetic-RL checkpoints whose base model is known show a similar pattern, with one pre-registered exception (Figure~\ref{fig:released}).
Of the two pre-registered primary models, RLVER-GRPO shows the effect under both judges ($+0.031$, $+0.056$), while MINT-4B, trained with a third-party empathy reward, does not under Qwen3.6 ($-0.004$) and only barely under Gemma-4 ($+0.026$ [$+0.002$, $+0.050$]); the pre-registered claim that both show it therefore did not hold.
MINT-1.7B, RLVER-PPO (with and without thinking) and Kardia-R1 show it under both judges (Kardia-R1: 0.30 $\to$ 0.64 under Qwen3.6); RLVER-GRPO with thinking is not usable (18--22\% of replies truncated). In total, 5 of 6 usable checkpoints show the effect under both judges.
No model endorses only seekers: where endorsement rises, the honesty effect is 0.44--1.7 times the seek effect.
These comparisons are observational, as each model differs from its base in data, reward and algorithm.

\subsection{Implications}
For the setting we test, the answer to the opening question is a threshold shift. The policy behaves like a learner that cannot identify the user's type and so optimises the average, as hidden-context theory \citep{siththaranjan2024distributional} and the mean-reward-gap account of agreement drift \citep{shapira2026agreement} would lead one to expect; the average takes the form of one endorsement criterion for everyone, placed by the share of users who reward endorsement, while discrimination between at-fault and not-at-fault posters is left intact. Stating the goal in words moves endorsement to the seekers but does not tie it more closely to fault.
Three practical consequences follow.
(1) The user mix is a training hyperparameter with a dose-dependent effect on endorsement, monotone in share on average, and RLVER's default mix (24.6\%) is enough to produce it.
(2) A fix should control the criterion rather than teach discrimination, which is intact, for instance by auditing the user mix or adding users who reward challenge.
(3) Evaluations should split by fault and include an explicit honesty request, since warmth and agreement benchmarks do not see the shift.
The reward-side penalties on endorsement we tried were seed-dependent, so we do not offer them as a fix (Appendix~\ref{app:patch}).

\section*{Limitations}\label{sec:limits}
We pose the question for RL from user reactions in general but answer it in one setting: empathetic RL against LLM-simulated users, evaluated on interpersonal-fault judgements; whether real users, other reward signals or other domains produce the same criterion shift is untested. Due to compute constraints, our controlled experiments use one policy family (Qwen2.5-7B, with a Llama-3.1-8B replication of one arm) and one user-simulator family (Qwen3.6). The effect depends on a simulator that rewards validation-seekers' preferred replies more than other replies; whether and how strongly other simulators do so is not established here.
Endorsement labels come from LLM judges, validated against two human annotators on 150 replies; Qwen3.6 is both the simulator and one judge, and the Gemma-4 judge and the human annotation serve as independent checks.
Evaluation is single-turn, on English posts, with requests in our own wording.
Several effects vary by seed; we report seeds individually and pool only where stated.
The released-model comparison is observational.
We do not offer a deployable mitigation.

\section*{Ethics statement}
The evaluation posts come from ELEPHANT's AITA sets \citep{cheng2025elephant}, which originate from Reddit; we use them for research evaluation only and redistribute no post text (our code refers to posts by dataset row). All models are used under the terms of their public releases (Apache-2.0, MIT and the Llama~3.1 Community License for the base and judge models). The work studies a failure mode of dialogue models trained for emotional support; we claim nothing about model feelings or intentions. Two annotators labelled model replies: university students recruited on campus, informed that the labels would be used for research, and paid 80 CNY per hour, above the local minimum wage. The task involved reading model replies to public Reddit posts and collected no personal information.

\arxivonly{\section*{Acknowledgments}
We thank the two annotators for their careful labelling. This work was supported by the Science and Technology Innovation 2030 -- Brain Science and Brain-Inspired Intelligence Major Project (2025ZD0215701) and the Beijing Natural Science Foundation (L257019).
}

\bibliography{refs}

\appendix
\section{Reward-side patches}\label{app:patch}
Penalising endorsement for users whose goal is reflection or dialectic (goal still hidden) gives a direction that varies by seed: lower, higher, and in the third seed lower in distribution but higher on out-of-distribution honesty requests ($+0.095$ Qwen3.6).
A first version of the fault-conditioned penalty collapsed; a second form that cannot be escaped by changing the reply removes endorsement in one seed, partly in one and not in a third, while protecting warranted validation better than an unconditional penalty ($+0.072$ [$+0.036$, $+0.108$] Qwen3.6).
We tried only our own variants; we did not reproduce \citeauthor{shapira2026agreement}'s exact penalty.

\Needspace{0.55\textheight}
\section{Training dynamics under the Gemma-4 judge}\label{app:dyn}
\begin{figure}[!h]
\includegraphics{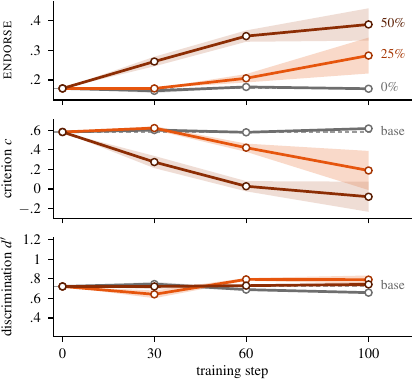}
\caption{As Figure~\ref{fig:dyn}, with labels from the Gemma-4 judge.}\label{fig:dyn-gemma}
\end{figure}

Figure~\ref{fig:dyn-gemma} repeats Figure~\ref{fig:dyn} with labels from the Gemma-4 judge.

\end{document}